\documentclass[conference]{IEEEtran}
\IEEEoverridecommandlockouts
\usepackage{cite}
\usepackage{amssymb}
\usepackage{amsmath,amsfonts}
\usepackage{graphicx}
\usepackage{textcomp}
\usepackage{xcolor}
\usepackage{stfloats}
\usepackage{algorithmic}
\usepackage{comment}
\usepackage{url}

\def\BibTeX{{\rm B\kern-.05em{\sc i\kern-.025em b}\kern-.08em
    T\kern-.1667em\lower.7ex\hbox{E}\kern-.125emX}}

\usepackage{fancyhdr}
\begin{document}

\title{Bias Reducing Multitask Learning on Mental Health Prediction\\
{\footnotesize \textsuperscript{}} 
\thanks{This work was supported by NSF \#1840167 and \#2047296.}
}

\author{\IEEEauthorblockN{Khadija Zanna, Kusha Sridhar, Han Yu, Akane Sano}
\IEEEauthorblockA{\textit{Department of Electrical and Computer Engineering} \\
\textit{Rice University}\\
Houston, USA \\
(khzanna, kh82, hy29, Akane.Sano)@rice.edu }




%
}

\maketitle

\thispagestyle{fancy}

\begin{abstract}
There has been an increase in research in developing machine learning models for mental health detection or prediction in recent years due to increased mental health issues in society. Effective use of mental health prediction or detection models can help mental health practitioners re-define mental illnesses more objectively than currently done, and identify illnesses at an earlier stage when interventions may be more effective. However, there is still a lack of standard in evaluating bias in such machine learning models in the field, which leads to challenges in providing reliable predictions and in addressing disparities. This lack of standards persists due to factors such as technical difficulties, complexities of high dimensional clinical health data, etc., which are especially true for physiological signals. This along with prior evidence of relations between some physiological signals with certain demographic identities restates the importance of exploring bias in mental health prediction models that utilize physiological signals. In this work, we aim to perform a fairness analysis and implement a multi-task learning based bias mitigation method on anxiety prediction models using ECG data. Our method is based on the idea of epistemic uncertainty and its relationship with model weights and feature space representation. Our analysis showed that our anxiety prediction base model introduced some bias with regards to age, income, ethnicity, and whether a participant is born in the U.S. or not, and our bias mitigation method performed better at reducing the bias in the model, when compared to the reweighting mitigation technique. Our analysis on feature importance also helped identify relationships between heart rate variability and multiple demographic groupings.

\end{abstract}


\section{Introduction and Background}

Irrespective of the advancements that machine learning has made possible in several fields such as language technologies, computer vision and medical applications, negative bias is often embedded in the essence of machine learning algorithms. Negative bias is an erroneous assumption made by an algorithm, that is systemically prejudiced against certain groups of people. Negative biases can be encoded in algorithms due to a number of factors, the first being imbalance in the representation of different population categories in the training data. If certain demographics are lacking from the sample data, models trained on this data often do not generalize when applied to new data that contains those missing demographics \cite{gorrostieta2019gender}. The second factor that could introduce negative bias in machine learning algorithms is biased human labeling. This is due to the fact that data that are fed into models, especially supervised or semi-supervised models which are widely used in various jurisdictions, are manually labeled by humans who are inherently biased. These models ultimately reflect people's impressions and sustain or further magnify bias from the labeled data \cite{vallor2017artificial}. Training and labeled data aside, there is still risk of introducing bias in the functional form of a model through features and modeling techniques \cite{gorrostieta2019gender}.

Due to the expanding popularity of machine learning and the inherent biases that come with it, there has been an increased focus on bias and fairness in the field. There are several works on how to accurately define and measure fairness in systems\cite{mehrabi2019survey,oneto2020fairness,srivastava2019mathematical}, how to analyze and mitigate bias using various techniques \cite{yan2020mitigating,oneto2019taking,gorrostieta2019gender}, and a few works that assess the trade-offs between fairness and accuracy in these models \cite{valdivia2021fair, rodolfa2021empirical}. 

Mental health poses a significant challenge for an individual's well-being, and it is estimated that 792 million people lived with a mental health disorder in 2017 \cite{owidmentalhealth}. This is slightly more than one in ten people globally (10.7\%). Rising statistics like this has led to an increase in research on mental health, including mental health and well-being prediction using physiological signals over the past couple of years. Several authors research on predicting stress levels, and various mental health conditions using data collected both in clinical settings and in the wild \cite{yu2020passive,xia2018physiological,dai2021comparing,albertetti2020stress,su2014emotiono+,swati2022early}.

With the rise in popularity of this field of research and its widespread applications in psychiatry and psychology, the need for effective bias mitigation techniques has become apparent, especially with the sensitive nature of physiological data. Previous research that explored emotional responses captured via physiological signals found some relations between blood pressure, electrodermal activity and race, and blood pressure and gender \cite{zanna2021clustering}. These findings re-iterate the importance of exploring possible bias in mental health prediction models that utilize physiological signals.

Despite several relatively successful bias mitigation efforts in general machine learning literature, when it comes to the field of mental health prediction and emotion recognition, there is still a lack of standards in methods for reducing bias. This ultimately leads to many challenges in providing reliable predictions and in addressing disparities \cite{park2021comparison}. This lack of standards persists due to factors such as technical difficulties in regard to data collection, complexities of high dimensional clinical health data, lack of knowledge of underlying causal structures, and challenges to algorithm evaluation \cite{mccradden2020ethical,rajkomar2018ensuring}. 

Only a few works to date have explored methods to reduce bias in mental health prediction models, however there has been research on bias analysis and  mitigation in emotion recognition models mostly on the facial recognition aspect of computer vision. Majority of these works state non-representative data ratio as a major cause of bias in facial recognition models \cite{howard2017addressing, xu2020investigating, panda2018contemplating, kara2021towards, chen2021towards}, and gender, age and skin-tone to be some of attributes data is unbalanced by \cite{menezes2021bias, kara2021towards, domnich2021responsible}. The methods these papers have introduced are often data and model-driven, making them non-transferable to emotion detection using other types of data \cite{xu2020investigating}. There have also been a few works that consider bias in emotion recognition using speech data \cite{gorrostieta2019gender,jalal2020removing}. They mention gender as a leading demographic that is a source of bias in models \cite{gorrostieta2019gender}. Other works conducted analyses on emotions in general healthcare procedures, and found that emotional responses in medical practices are heavily culturally mediated with both individual factors like gender, age, occupation, and social factors like food habits, availability, etc. coming into play \cite{casacuberta2022biases}.

Park et al. explored the performance of different methods to reduce bias for clinical prediction algorithms for postpartum depression \cite{park2021comparison}. They implemented three methods, reweighting, prejudice removal, and removal of the race label to logistic regression, random forest, and extreme gradient boosting models, to mitigate bias based on race. They found that reweighing improves their chosen fairness metrics without compromising accuracy, prejudice remover performed less reliably, and omitting the race label made no significant difference to the fairness metric. Although this study provided promising results, there are some limitations that come with it. First, the authors conducted their experiments based on data collected in a clinical setting, and these results might not hold true on data from the wild. They also implemented these methods on well-known interpretable, and often used models in fairness literature \cite{valdivia2021fair}, and only analyzed bias based on race without exploring other factors. To counter some of these limitations, we will be testing our method on a Long Short term memory (LSTM)-based anxiety prediction model using physiological data collected in the wild. The reason we are using anxiety to assess fairness and our bias mitigation method is due to the fact that most scientific work addressing anxiety and its disorders has thus far been conducted among European Americans. This has created striking gaps of inequalities in anxiety disorder research and practice \cite{zvolensky2017disparities}, and with this work, we aim to aid in closing some of this gap.

One area of interest in the field of fairness in general is the trade-off between accuracy and fairness in models \cite{valdivia2021fair,oneto2019taking}. It is often the case that bias is first introduced to models from the data, and that means using sensitive information in the functional form of the model will improve prediction accuracy \cite{oneto2019taking}. However, it is well known that in some jurisdictions, using different models either explicitly or implicitly for different protected groups is not allowed \cite{oneto2019taking}. In this paper, we develop a technique that enables us to optimize accuracy and fairness while improving interpretability of the model, without explicitly using any of the sensitive information in the functional form of our model. We propose the use of Multitask Learning (MTL), which has been proven to improve interpretablity in models that use multidimensional health data \cite{zhang2021survey}, and has been used in previous research to optimize both accuracy and fairness \cite{oneto2019taking}, along with Monte Carlo dropout to utilize model uncertainty to improve fairness without sacrificing either computational complexity or accuracy. 

Our contribution can be summarized as 

(i) analyses of bias based on different demographic information on data collected in the wild for anxiety prediction, 

(ii) development of an MTL-based bias mitigation technique to optimize fairness while preserving accuracy and improving interpretability,

(iii) evaluation of the proposed method on LSTM-based models and a large scale of public biobehavioral dataset with various demographics, and

(iv) comparison of our method against a conventional method, reweighting introduced by \cite{park2021comparison} for both accuracy and fairness measures.


\section{Data and Methods}

\subsection{Dataset}

In our experiments, we used a part of the TILES dataset, physiological data collected by Mundnich et al. \cite{mundnich2020tiles} from over 200 hospital workers. We used 10 weeks worth of electrocardiogram (ECG) data collected with the OMSignal smart garment (15-second long ECG signal in 250 Hz every 5 minutes). We extracted 25 frequency and time-domain ECG features, such as the statistical characteristics of the peak-to-peak intervals and the energy of the signal in various frequency bands. Table \ref{tab:features} shows the full list of the extracted tables, and the definition of each features can be referred to \cite{hrvanalysis}. We made use of self-reported anxiety levels which were measured using the State Trait Anxiety Inventory giving a value in the range 20 to 80 for each participant. We further binarized these scores using personalized z-score. The original released data contains anxiety labels that were reported in a 5-point scale by subjects. Following the setting in \cite{gaballah2021context}, we calculated the z-score of labels for each participant separately, and marked labels below the personalized average as negative (0) and labels above the mean as positive (1). We used 2 hours of the data (5 minutes x 24 steps) to infer upcoming anxiety labels. Under this scenario, we have in total 920 samples with 506 negative samples and 414 positive samples.

We used gender, age, race, income, shift, ethnicity, born in the US, English as a native language, and work hours as the demographic data (referred to as the 9 protected labels in this paper) to test our model fairness on. These data were collected via surveys administered to the participants during the study. We binary-encoded each of the protected labels to assign them as privileged and unprivileged classes to effectively analyze them using our chosen fairness metrics. We chose the class with the higher number of participants as the privileged class (denoted by 1), and the other with less participants as the unprivileged class (denoted by 0). Figure \ref{fig1} below shows the distribution of the participants in our data based on the different demographic groups.
We split the data into 75\% training and 25\% testing sets to fit into the model. 

\begin{table}[h]
\centering
\caption{List of extracted features}

\begin{tabular}{c|c}
\hline
Category                 & Features                                                                                                                                                                                            \\ \hline
Time Domain      & \begin{tabular}[c]{@{}c@{}}mean\_nni, sdnn, sdsd, nni\_50, pnni\_50, \\ nni\_20, pnni\_20,rmssd, median\_nni, range\_nni, \\ cvsd, cvnni, mean\_hr, max\_hr, min\_hr, std\_hr\end{tabular} \\ \hline
Frequency Domain & lf, hf, lf\_hf\_ratio, lfnu, hfnu, total\_power, vlf                                                                                                                                       \\ \hline
Other           & cardiac sympathetic index, cardiac vagal index                                                                                                                                             \\ \hline
\end{tabular}
\label{tab:features}
\end{table}


\begin{figure}[htbp]
\centerline{\includegraphics[width=\columnwidth]{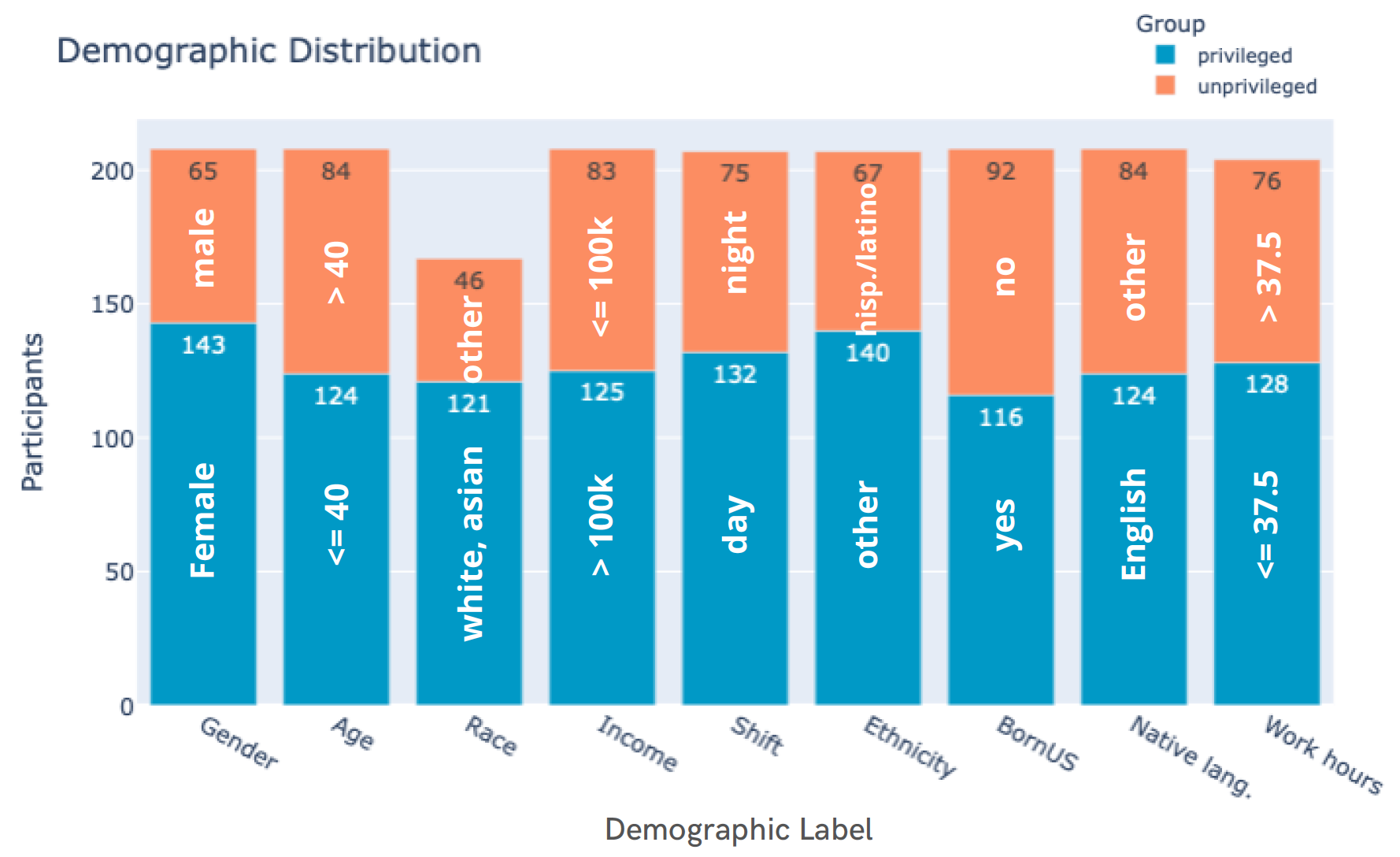}}
\caption{Distributions of Demographic Data in the TILES dataset}
\label{fig1}
\end{figure}

\subsection{Fairness Terminology} \label{ssec:fair_terms}

In this section, we introduce and define some of the concepts and terminologies generally used in algorithmic fairness research that are relevant to this paper.

\begin{itemize}
\item \textbf{Protected (sensitive) label}: An attribute that partitions a population into groups whose outcomes should have parity (such as race, gender, income, etc.).
\item \textbf{Privileged class}: A protected label value indicating a group that is at an advantage.
\item \textbf{Unprivileged class}: A protected label value indicating a group that is at a disadvantage.
\item \textbf{Disparate impact ratio (DIR)}: This is a fairness metric which is the ratio of positive outcomes (anxiety=1) in the unprivileged class, divided by the ratio of positive outcomes in the privileged class, as shown in equation \ref{eq:disp_imp}. It is the measure of how different outcomes are for different groups, based on the results of a model \cite{feldman2015certifying}.
In this paper, we assume an acceptable lower bound of 0.8, and a higher bound of 1.2, with 1 being the the ideal score. 

\begin{equation}
    \text{DIR =} \frac{Pr(Y=1|D=unprivileged)} {Pr(Y=1|D=privileged)}
\label{eq:disp_imp}
\end{equation}

\item \textbf{Equalized odds}: This fairness metric enforces that the model correctly identify the positive outcome at equal rates across groups, and misclassify the positive outcome at equal rates across groups (creating the same proportion of True Positives and False Positives across groups). In the context of this paper, we choose to compare false negative instead of true positive rates between the privileged and unprivileged classes because a negative outcome (anxiety=0) is more desirable in our study.


\end{itemize}

\subsection{Multi-task Learning Based Bias Mitigation Method}

Our proposed methodology is based on the premise of epistemic uncertainty in Bayesian uncertainty estimation. Epistemic uncertainty refers to uncertainty in the structure and parameters of a model caused by a lack of knowledge, which has also been proven to correlate with model weights \cite{li2021deep, wang2021hybrid}. We hypothesize that when the model is most uncertain about the protected label, the weights of that model will lack knowledge at certain regions of the feature space related to the protected label, and therefore using these weights for anxiety prediction will minimize the influence of that protected label on the final anxiety prediction. Our proposed method will in turn minimize any bias that might be introduced to the model through imbalances in the data based on the protected label.

\begin{figure*}[htbp]
\centerline{\includegraphics[width=\textwidth]{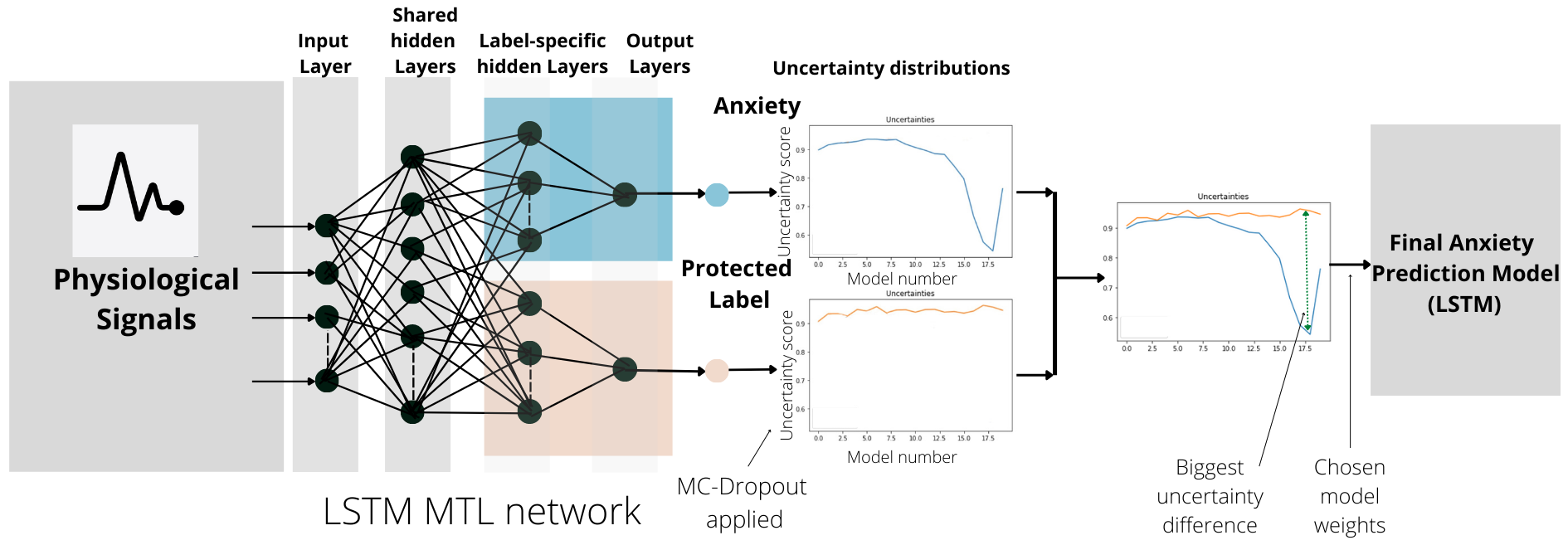}}
\caption{Process Diagram}
\label{fig2}
\end{figure*}

Figure \ref{fig2} presents a visual diagram of our proposed method. We utilize multi-task learning to predict anxiety and one of the protected labels (e.g gender). We save the model periodically in our experiments, and implement Monte Carlo dropout which allows us to get uncertainty estimations of these saved models. Modeling uncertainty with Monte Carlo dropout works by running multiple forward passes through the model with a different dropout mask every time. Given a trained neural network model with dropout \(f_{nn}\), to derive uncertainty for one sample \(x\), we collect the predictions of T inferences with different dropout masks. Here, \(f_{nn}^{d_i}\) represents the model with dropout mask \(d_i\). So we obtain a sample of the possible model outputs for sample \(x\) as  \(f_{nn} ^{d_0} (x),...,f_{nn} ^{d_T} (x)\).


By computing the average and variance of this sample, we obtain an ensemble prediction, which is the mean of the models posterior distribution for this sample and an estimate of the uncertainty of the model regarding \(x\).

\begin{equation}
\label{eq:eq-mean}
    \text{predictive posterior mean: } p = \frac{1}{T} \sum_{i=1}^{T} f_{nn} ^{d_i} (x)
\end{equation}

\begin{equation}
\label{eq:uncert}
    \text{uncertainty: } c = \frac{1}{T} \sum_{i=1}^{T}[f_{nn} ^{d_i} (x) - p]^{2}
\end{equation}

More information on this can be found in the original publication by Gal et. al \cite{gal2016dropout}.

From the distribution of Monte Carlo predictions obtained, we select the model prediction where the difference in uncertainties between anxiety and the protected label is highest, and then extract the parameters of the model at this desired point, and utilize them in our final anxiety prediction model.

\section{Experiments} 

\subsection{Analysis of Base Model}

To start off our experiments, we calculated the disparate impact ratio of the original dataset for a combination of every protected label and anxiety, to understand if the data itself was biased in terms of the anxiety labels. Next, we ran a basic LSTM model for anxiety prediction and analyzed our results for fairness by calculating the disparate impact ratio and equalized odds for our prediction against all 9 protected labels. We conducted these analyses to determine whether and where the model was introducing bias, and ensure that we only apply our bias mitigation method on aspects of the data that are actually biased. 

After performing fairness analysis on the base model, we selected the protected labels by which our model showed most bias, and tested our method on.


\subsection{Method Implementation}

\begin{figure}[htbp]
\centerline{\includegraphics[width=5cm, height=6.5cm]{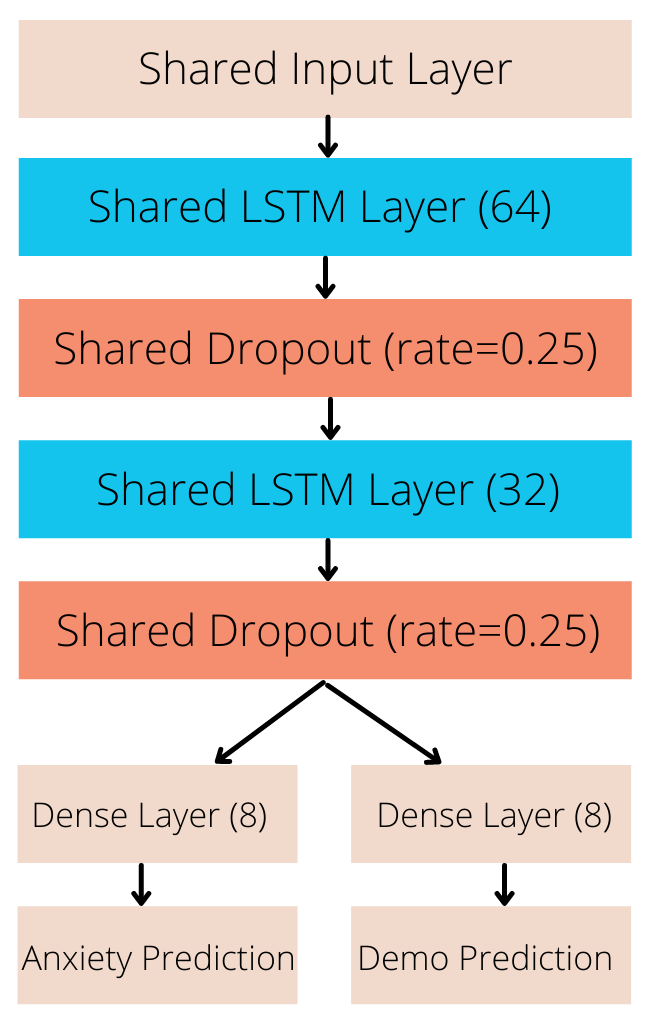}}
\caption{Structure of Multi-talk Model}
\label{fig3}
\end{figure}

To implement our method, we ran the MTL model with the structure shown in Figure \ref{fig3} on a combination of anxiety plus each of the protected labels that showed bias in the base model, and we assigned loss weights of the ratio 4.5:0.5, 4.5 for anxiety and 0.5 for the demographic label. We used this ratio to ensure that anxiety prediction is prioritized by the model, making sure that its uncertainty score is kept lower than the demographic prediction. We arrived at this combination by experimenting with a few different combinations.

We ran these MTL networks for 100 epochs each, and saved the weights at every 5 epochs, making it a total of 20 weight combinations, and we determined these parameters using a grid search. With this, we used Monte Carlo dropout, and ran some analysis on the Monte Carlo outputs to calculate the uncertainties of each of the saved weight combinations. We computed the uncertainties by calculating the sample variance of the different forward passes using the equation \ref{eq:uncert}.

We plotted the uncertainties and identified the model that showed the biggest difference in uncertainty scores between anxiety and the demographic prediction. 

\subsection{Saliency Maps}
To gain a better understanding of how the features influence our anxiety prediction and the different protected labels, and how our proposed method affects the models, we utilized a saliency map technique \cite{simonyan2013deep} to visualize the importance of model weights on our input time axis and features. According to the predicted class $c$, for example, the decision-making process of the model can be represented as $S_{c}(I) = w_{c}^{T}I + b_c$, where $I ,S_c(I), w, b$ are the model input, output, weights and bias, respectively. The essential of this method is to calculate the model weights on the input layer by gradients, e.g., $w = \frac{\partial S_c}{\partial I}|_{I_0}$. The calculated $w$ represents the model saliency regrading to the input layer. In this study, we fetched the saliency maps for all the test samples and calculated the average saliency map to develop the intuition of important features and time steps in general.

\section{Results}

\subsection{Initial Analysis}

\begin{figure}[htbp]
\centerline{\includegraphics[width=\columnwidth]{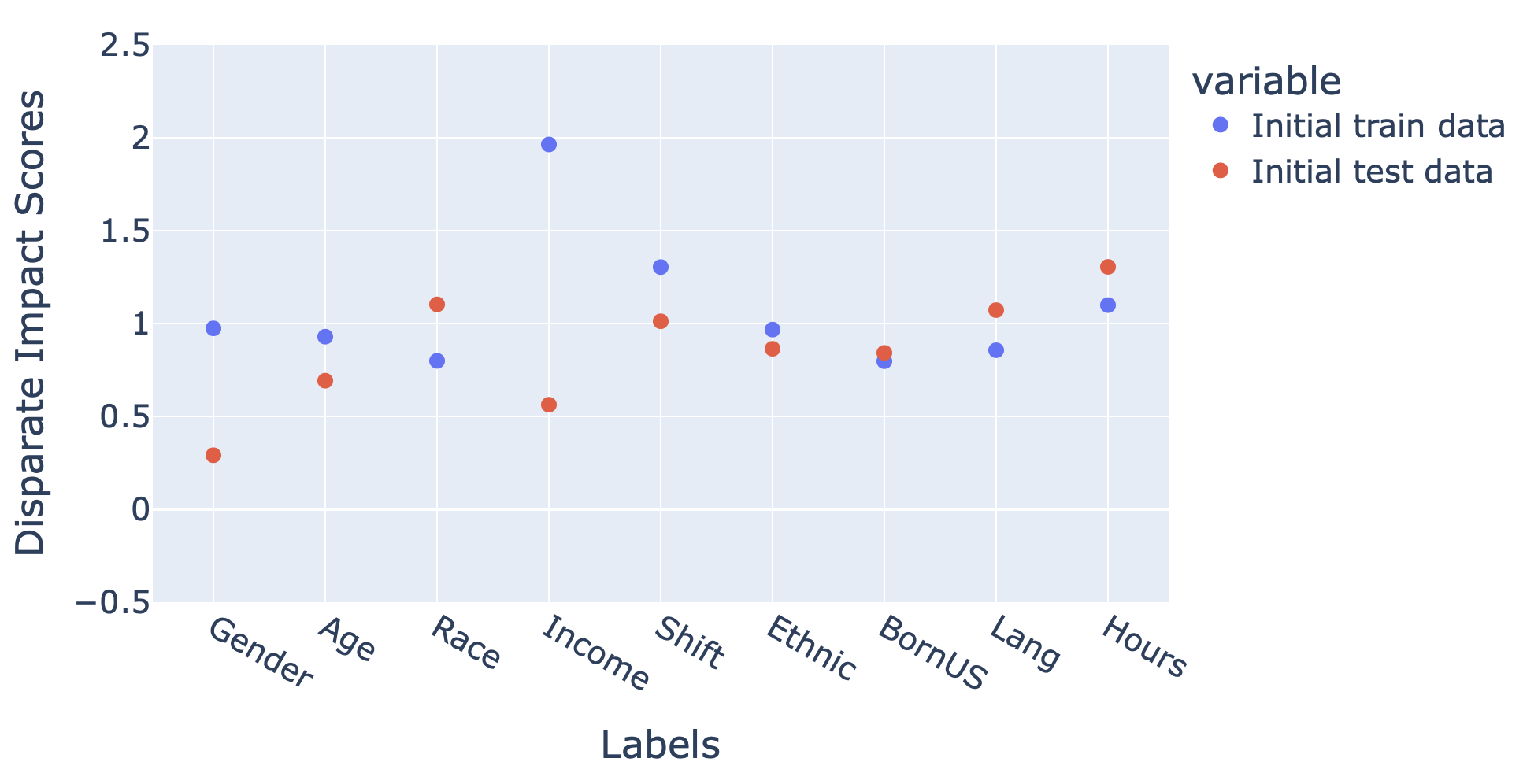}}
\caption{Disparate Impact Scores of Initial Data}
\label{fig4}
\end{figure}

After analyzing the dataset before running any models, the TILES dataset appeared to be relatively balanced in terms of the disparate impact ratio. Figure~\ref{fig4} shows the distribution of these scores and their proximity to the ideal score of 1. For most of the protected labels, the scores for the training data were in the range of 0.8 to 1.2, with the exception of income and shift with scores of 1.96 and 1.3 respectively. As for the testing data, the scores for race, shift, ethnicity, bornUS, and lang were in the range, while scores for gender, age income are below, and hours were above the range. 

After running the base LSTM model, we obtained an accuracy score of 57.5\% and an F1 score of 0.487. The baseline random performance for anxiety prediction was 49\%, and the macro F1 score was 0.348.

A fairness analysis of the predictions showed us that the model does introduce some bias for some of the labels, specifically age, income, ethnicity, and bornUS, as shown in Tables \ref{tab:scores_comparison_age} to \ref{tab:scores_comparison_bornUS}. The scores for the other labels fall within the chosen range of 0.8 and 1.2. 

\begin{figure}[htbp]
\centerline{\includegraphics[width=0.95\columnwidth]{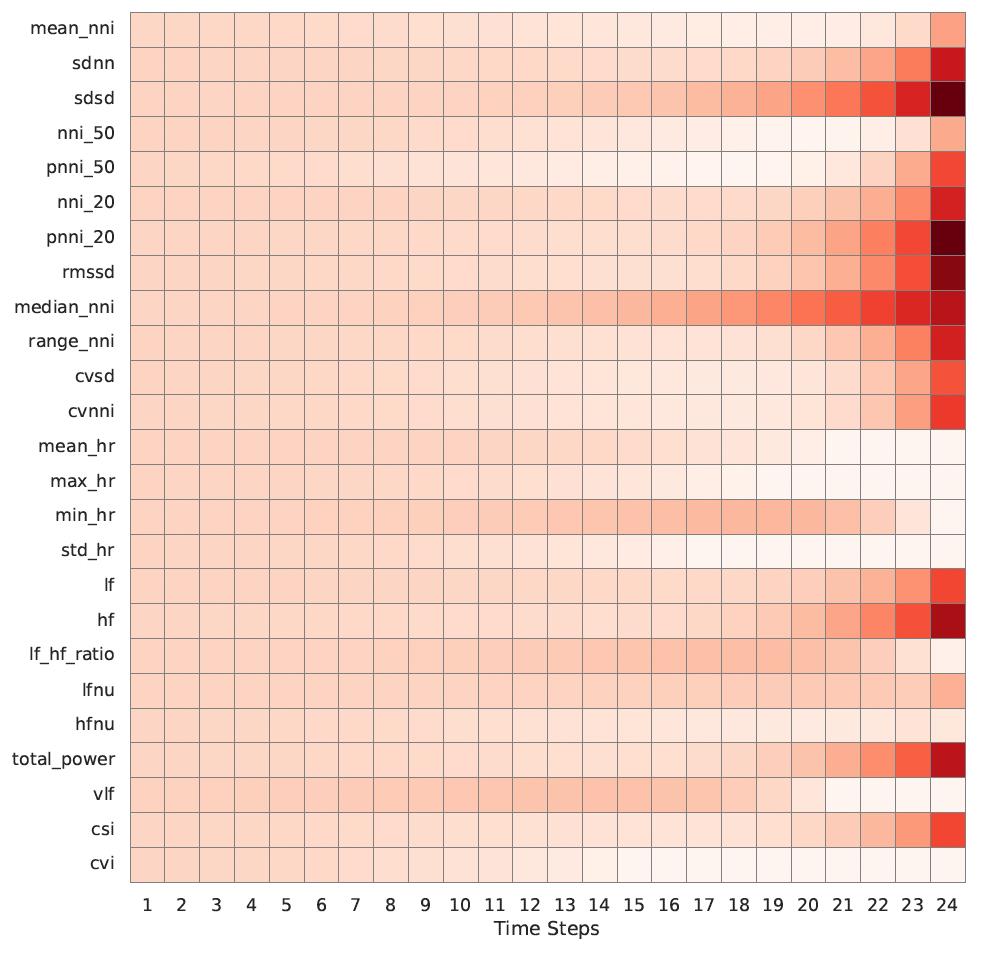}}
\caption{Saliency Map of Base Anxiety Prediction (5 mins x 24 steps). The Y-axis indicates the feature names, where the X-axis represents the time steps of the sequences. Each time step represents the each feature point in the input sequences, and the number of the time steps is equal to the length of input sequences.}
\label{fig5}
\end{figure}

Figure \ref{fig5} shows the saliency map of anxiety produced by the base LSTM model. The x-axis represents the different features in our data, and the y-axis represents the 24 timesteps fed into the LSTM model to infer upcoming anxiety labels. From this figure, we can see that feature importance is relatively evenly distributed, with the most importance given to sdsd and  pnni\_20, which are time-domain-based heart rate variability features. In general, heart rate features (mean\_hr, max\_hr, min\_hr, std\_hr) seem to have the least importance.

\subsection{Proposed Method}

Next, we implemented our method to predict anxiety while mitigating bias caused by age, income, ethnicity, and bornUS protected labels.

Tables \ref{tab:scores_comparison_age} to \ref{tab:scores_comparison_bornUS} show a comparison of performance and fairness scores for the base model, our method, and an implementation of the reweighting method on our data. Reweighting involves applying appropriate weights to different tuples in the training dataset to make it discrimination free with respect to the protected label \cite{park2021comparison}. These tables show that our method was able to improve all the fairness scores, giving an ideal disparate impact ratio score of 1 for, and 0 for differences in False Positive and False Negative rates. This comes at a performance cost, as both our method and reweighting were not able to preserve the accuracy of the base model and F1 scores of the base model.

\begin{figure*}[htbp]
\centerline{\includegraphics[width=\textwidth]{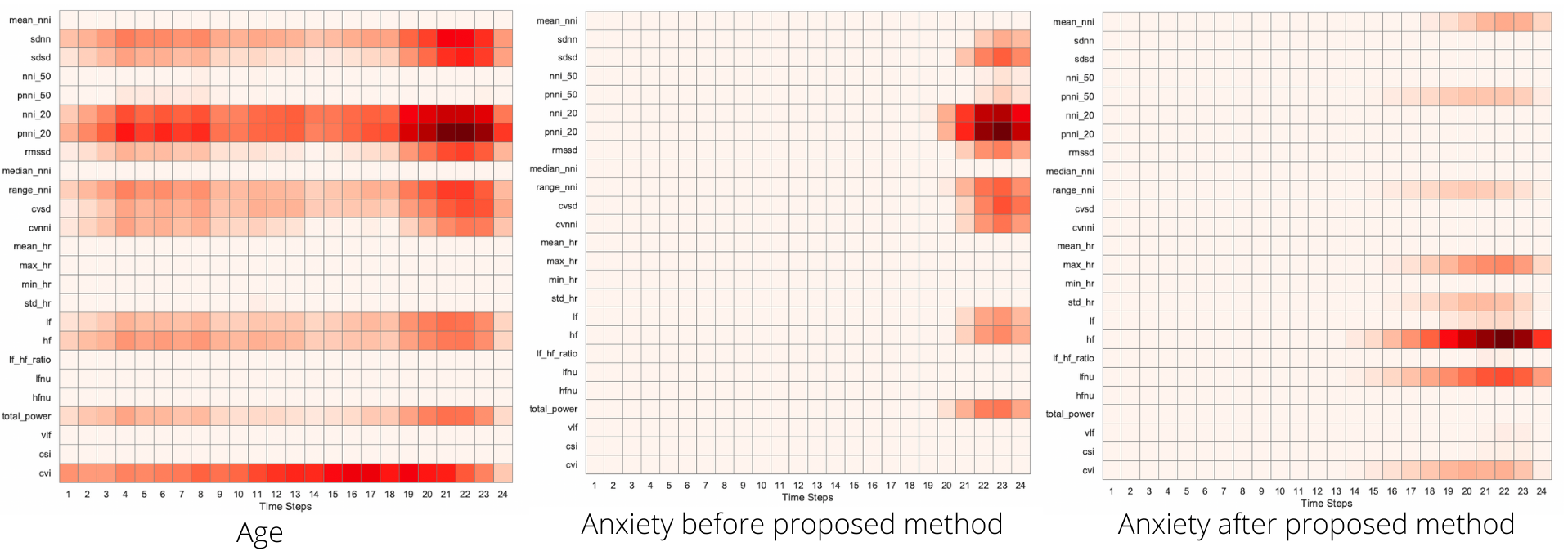}}
\caption{Saliency Maps of Age Experiments}
\label{fig_sal_age}
\end{figure*}

\begin{figure*}[htbp]
\centerline{\includegraphics[width=\textwidth]{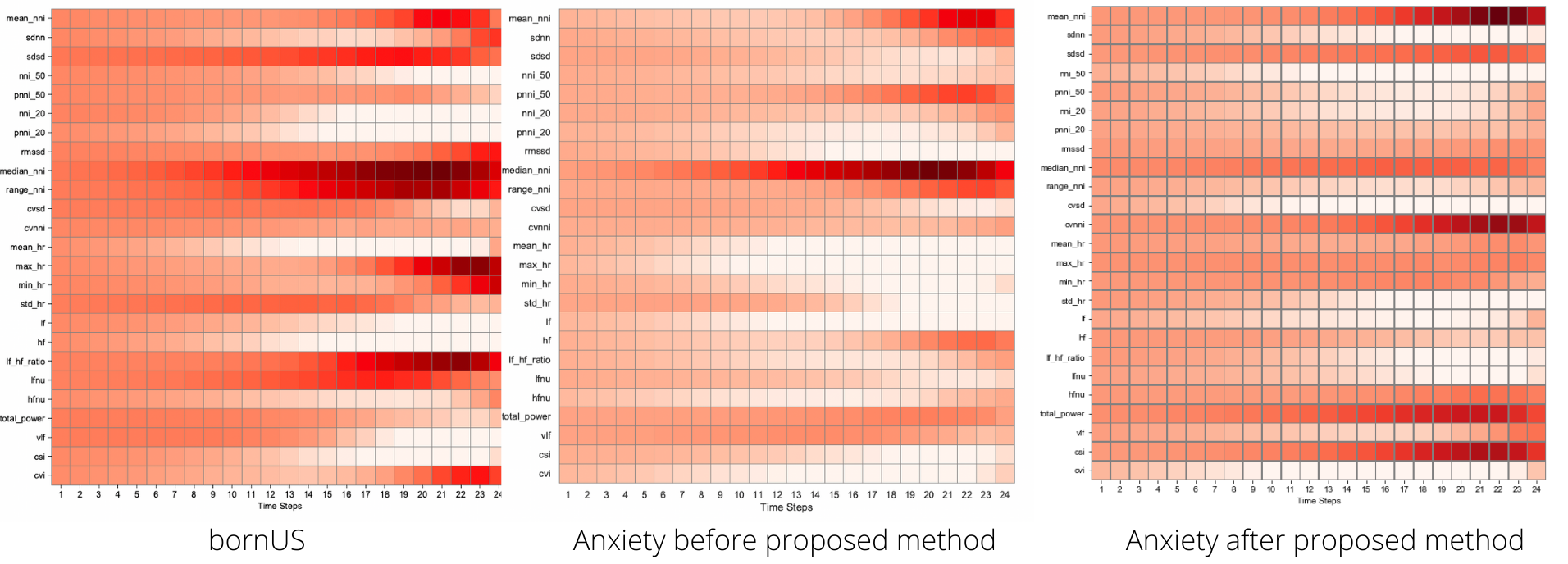}}
\caption{Saliency Maps of bornUS Experiments}
\label{fig_sal_bornUS}
\end{figure*}

Figures \ref{fig_sal_age} and \ref{fig_sal_bornUS} show the saliency maps for the age and bornUS experiments. Darker red color indicates more importance. The first map on each figure shows the feature importance on the demographic prediction when running the MTL model, the second shows that of anxiety from the MTL model before proposed method was implemented, and the third one for anxiety after method was fully implemented. Our method shifted weights from features that are important to the protected label to those that are less important. This ultimately decreased the importance of features associated with the protected label in the final anxiety prediction. For example, in Figure \ref{fig_sal_age}, the first map shows that features nni\_20 and pnni\_20 carried the most importance when it comes to the age label, and the second map shows that before our method implementation, they also appeared to be very important for anxiety prediction. After our method was implemented, it is shown in the third map that these features carried less weight and ultimately are less important for anxiety prediction, which according to our hypothesis, reduced the possibility of the model being biased based on age.

\begin{table}[htbp]
\centering
\label{tbl:scores_comparison_age}
\caption{Comparison of Models based on Age 
}
\resizebox{\columnwidth}{!}{  
\renewcommand{\arraystretch}{1.3}

\begin{tabular}{|l|l|l|l|}
\hline
\textbf{Metric} & \textbf{Base Model} & \textbf{Reweighting} & \textbf{Proposed Method}\\ \hline
Accuracy & 0.575 & 0.549 & 0.527  \\ \hline
F1 & 0.487 & 0.517 & 0.418 \\ \hline
DI Ratio & 0.682 & 0.648 & 1 \\ \hline
Diff in FN & 0.077  & 0.066 & 0 \\ \hline
Diff in FP & -0.077 & -0.059 & 0 \\ \hline
\end{tabular}}
\label{tab:scores_comparison_age}
\end{table}

\vspace{-5mm}

\begin{table}[htbp]
\centering
\label{tbl:scores_comparison_income}
\caption{Comparison of Models based on Income}
\resizebox{\columnwidth}{!}{  
\renewcommand{\arraystretch}{1.3}

\begin{tabular}{|l|l|l|l|}
\hline
\textbf{Metric} & \textbf{Base Model} & \textbf{Reweighting} & \textbf{Proposed Method} \\ \hline
Accuracy & 0.575 & 0.539 & 0.544  \\ \hline
F1 & 0.487 & 0.517 & 0.408 \\ \hline
DI Ratio & 1.389 & 0.997 & 1 \\ \hline
Diff in FN & -0.099 & -0.153 & 0 \\ \hline
Diff in FP & 0.055 & -0.114 & 0 \\ \hline
\end{tabular}}
\label{tab:scores_comparison_income}
\end{table}
\vspace{-5mm}


\begin{table}[htbp]
\centering
\label{tbl:scores_comparison_ethnicity}
\caption{Comparison of Models based on Ethnicity}
\resizebox{\columnwidth}{!}{  
\renewcommand{\arraystretch}{1.3}

\begin{tabular}{|l|l|l|l|}
\hline
\textbf{Metric} & \textbf{Base Model} & \textbf{Reweighting} & \textbf{Proposed Method} \\ \hline
Accuracy & 0.575 & 0.542 & 0.434 \\ \hline
F1 & 0.487 & 0.450 & 0.303 \\ \hline
DI Ratio & 1.24 & 1.329 & 1 \\ \hline
Diff in FN & 0.072 & -0.054 & 0 \\ \hline
Diff in FP & 0.128 & 0.054 & 0 \\ \hline
\end{tabular}}
\label{tab:scores_comparison_ethnicity}
\end{table}
\vspace{-5mm}


\begin{table}[htbp]
\centering
\label{tbl:scores_comparison_bornUS}
\caption{Comparison of Models based on "Born in the US"}
\resizebox{\columnwidth}{!}{  
\renewcommand{\arraystretch}{1.3}

\begin{tabular}{|l|l|l|l|}
\hline
\textbf{Metric} & \textbf{Base Model} & \textbf{Reweighting}  & \textbf{Proposed Method} \\ \hline
Accuracy & 0.575 & 0.525 &  0.504 \\ \hline
F1 & 0.487 & 0.490 &  0.388 \\ \hline
DI Ratio & 0.673 & 0.866 & 1 \\ \hline
Diff in FN & 0.110 & 0.094 & 0 \\ \hline
Diff in FP & -0.058 & 0.016 & 0 \\ \hline
\end{tabular}}
\label{tab:scores_comparison_bornUS}
\end{table}

\section{Discussion and conclusion}

Seeing the ever growing effects of mental health on people's lives, and the gap between advances made in mental health prediction research, and development of effective bias analysis and mitigation methods, the goal of our study is to develop an effective technique to identify and mitigate bias in mental health prediction models. To do this, we analyzed bias in a physiological dataset, and an anxiety prediction model, and introduced a multi-task learning-based bias mitigation method where necessary. This method is based on the hypothesis that when a model is most uncertain about a particular protected label, the weights of that model will lack knowledge at certain regions of the feature space related to that label. Which means that using the weights of that model to predict anxiety will minimize the influence of the protected label on anxiety prediction.  
Our analysis found that the TILES dataset on its own was imbalanced by number of participants based on gender, age, race, income, shift, ethnicity, number of work hours, and whether or not the participant was born in the US (bornUS), but it was not biased by these demographics when analyzed using the disparate impact ratio as a fairness metric. After running an LSTM model to predict anxiety, and analyzing the fairness in the prediction results, we discovered that the model introduced some bias by age, income, ethnicity, and bornUS. 

We implemented our method to mitigate bias by each of these protected labels, and compared our results to a standard reweighting bias mitigation technique. Our results show that our method overall did better than reweighting on the fairness metric scores, but was not able to preserve the anxiety model performance. There was an average of 7\% drop in accuracy, across all four experiments (age, income, ethnicity, bornUS), with the experiment on ethnicity causing the most depreciation by 14\%.  

To better understand the effect of different features on prediction, and how they are related to anxiety, and the different protected labels, we utilized a saliency map technique to visualize the importance of model weights on different prediction tasks. We found feature importance to be relatively evenly distributed for anxiety prediction but most correlated with 2 time-domain-based heart rate variability features (sdsd and pnni\_20). Heart rate variability is the fluctuation of heart period over time, commonly measured by ECG, and is an important marker of psychological well-being \cite{chalmers2014anxiety}. One study found anxiety disorders to correspond with lower heart rate variability (HRV) \cite{chalmers2014anxiety}, confirming why these features carry the most weight for anxiety prediction.

The saliency maps for the protected labels also gave us some interesting insight. We found that age correlated most with time domain-based HRV features (pnni\_20, nni\_20, sdsd, sdnn, cvi). Research has shown connections between HRV and a number of demographic factors such as age, race, ethnicity and gender or sex \cite{zhang2007effect, umetani1998twenty,choi2006age}. This was confirmed by the high correlation between HRV features with bornUS and ethnicity as well. Income on the other hand, correlated with both frequency domain and time domain-based HRV features. Seeing as the considered protected labels and anxiety have high correlations with certain HRV features, it is easy to see how bias can be introduced to the model based on these protected labels. 

The saliency maps also showed how our method shifted importance from features that highly correlate with a particular protected label, in order to reduce it's effect on the final anxiety prediction. Having stated that, we attribute our method's lack of ability to preserve accuracy and F1 scores to the fact that both anxiety and the protected labels have similar important features, which means that shifting weights from these features will undoubtedly have a significant effect on the model's ability to form a pattern for anxiety, ultimately affecting it's prediction. This method will benefit from further testing using datasets with more varying label-feature relationships.  

In conclusion, it is important to acknowledge that data sources represent just one aspect of bias, which can be introduced through a model and certain technical metrics as well. The purpose of prediction algorithms is to influence the clinical decision-making process, and the biases of those developing and using them would have greater impact on whether they ultimately perpetuate inequality. For this reason, it is imperative to take into consideration, the impact of both human behavior and technical details when developing algorithms that make critical decisions such as those on mental health, especially in clinical settings. 

Working with deep learning models for our prediction task, we have encountered a few limitations during the course of our study. First is the computational complexity of using monte carlo dropout. Using this method along with the aspect of saving and loading model weights results in a higher than normal computational cost. In the future, we aim to test our bias mitigation method on other datasets both physiological and non-physiological, and explore different ways to better preserve accuracy with it. We will explore developing a custom loss function that will utilize the relationship between uncertainty and weights to produce better performance.

\bibliographystyle{ieee}
\bibliography{references}

\begin{thebibliography}{10}\itemsep=-1pt

\bibitem{albertetti2020stress}
F.~Albertetti, A.~Simalastar, and A.~Rizzotti-Kaddouri.
\newblock Stress detection with deep learning approaches using physiological
  signals.
\newblock In {\em International Conference on IoT Technologies for HealthCare},
  pages 95--111. Springer, 2020.

\bibitem{casacuberta2022biases}
D.~Casacuberta and J.~Vallverd{\'u}.
\newblock Biases in assigning emotions in patients due to multicultural issues.
\newblock In {\em Handbook of Artificial Intelligence in Healthcare}, pages
  215--228. Springer, 2022.

\bibitem{chalmers2014anxiety}
J.~A. Chalmers, D.~S. Quintana, M.~J. Abbott, A.~H. Kemp, et~al.
\newblock Anxiety disorders are associated with reduced heart rate variability:
  a meta-analysis.
\newblock {\em Frontiers in psychiatry}, 5:80, 2014.

\bibitem{hrvanalysis}
R.~Champseix.
\newblock Heart rate variability analysis.
\newblock \url{https://github.com/Aura-healthcare/hrv-analysis}, 2018.

\bibitem{chen2021towards}
Y.~Chen, X.~Yang, T.-J. Cham, and J.~Cai.
\newblock Towards unbiased visual emotion recognition via causal intervention.
\newblock {\em arXiv preprint arXiv:2107.12096}, 2021.

\bibitem{choi2006age}
J.-B. Choi, S.~Hong, R.~Nelesen, W.~A. Bardwell, L.~Natarajan, C.~Schubert, and
  J.~E. Dimsdale.
\newblock Age and ethnicity differences in short-term heart-rate variability.
\newblock {\em Psychosomatic medicine}, 68(3):421--426, 2006.

\bibitem{dai2021comparing}
R.~Dai, C.~Lu, L.~Yun, E.~Lenze, M.~Avidan, and T.~Kannampallil.
\newblock Comparing stress prediction models using smartwatch physiological
  signals and participant self-reports.
\newblock {\em Computer Methods and Programs in Biomedicine}, 208:106207, 2021.

\bibitem{domnich2021responsible}
A.~Domnich and G.~Anbarjafari.
\newblock Responsible ai: Gender bias assessment in emotion recognition.
\newblock {\em arXiv preprint arXiv:2103.11436}, 2021.

\bibitem{feldman2015certifying}
M.~Feldman, S.~A. Friedler, J.~Moeller, C.~Scheidegger, and
  S.~Venkatasubramanian.
\newblock Certifying and removing disparate impact.
\newblock In {\em proceedings of the 21th ACM SIGKDD international conference
  on knowledge discovery and data mining}, pages 259--268, 2015.

\bibitem{gaballah2021context}
A.~Gaballah, A.~Tiwari, S.~Narayanan, and T.~H. Falk.
\newblock Context-aware speech stress detection in hospital workers using
  bi-lstm classifiers.
\newblock In {\em ICASSP 2021-2021 IEEE International Conference on Acoustics,
  Speech and Signal Processing (ICASSP)}, pages 8348--8352. IEEE, 2021.

\bibitem{gal2016dropout}
Y.~Gal and Z.~Ghahramani.
\newblock Dropout as a bayesian approximation: Representing model uncertainty
  in deep learning.
\newblock In {\em international conference on machine learning}, pages
  1050--1059. PMLR, 2016.

\bibitem{gorrostieta2019gender}
C.~Gorrostieta, R.~Lotfian, K.~Taylor, R.~Brutti, and J.~Kane.
\newblock Gender de-biasing in speech emotion recognition.
\newblock In {\em INTERSPEECH}, pages 2823--2827, 2019.

\bibitem{howard2017addressing}
A.~Howard, C.~Zhang, and E.~Horvitz.
\newblock Addressing bias in machine learning algorithms: A pilot study on
  emotion recognition for intelligent systems.
\newblock In {\em 2017 IEEE Workshop on Advanced Robotics and its Social
  Impacts (ARSO)}, pages 1--7. IEEE, 2017.

\bibitem{jalal2020removing}
M.~A. Jalal, R.~Milner, T.~Hain, and R.~K. Moore.
\newblock Removing bias with residual mixture of multi-view attention for
  speech emotion recognition.
\newblock In {\em Interspeech 2020}, pages 4084--4088. ISCA-International
  Speech Communication Association, 2020.

\bibitem{kara2021towards}
O.~Kara, N.~Churamani, and H.~Gunes.
\newblock Towards fair affective robotics: Continual learning for mitigating
  bias in facial expression and action unit recognition.
\newblock {\em arXiv preprint arXiv:2103.09233}, 2021.

\bibitem{li2021deep}
Y.~Li, S.~Rao, A.~Hassaine, R.~Ramakrishnan, D.~Canoy, G.~Salimi-Khorshidi,
  M.~Mamouei, T.~Lukasiewicz, and K.~Rahimi.
\newblock Deep bayesian gaussian processes for uncertainty estimation in
  electronic health records.
\newblock {\em Scientific reports}, 11(1):1--13, 2021.

\bibitem{mccradden2020ethical}
M.~D. McCradden, S.~Joshi, M.~Mazwi, and J.~A. Anderson.
\newblock Ethical limitations of algorithmic fairness solutions in health care
  machine learning.
\newblock {\em The Lancet Digital Health}, 2(5):e221--e223, 2020.

\bibitem{mehrabi2019survey}
N.~Mehrabi, F.~Morstatter, N.~Saxena, K.~Lerman, and A.~Galstyan.
\newblock A survey on bias and fairness in machine learning. arxiv 2019.
\newblock {\em arXiv preprint arXiv:1908.09635}, 2019.

\bibitem{menezes2021bias}
H.~F. Menezes, A.~S. Ferreira, E.~T. Pereira, and H.~M. Gomes.
\newblock Bias and fairness in face detection.
\newblock In {\em 2021 34th SIBGRAPI Conference on Graphics, Patterns and
  Images (SIBGRAPI)}, pages 247--254. IEEE, 2021.

\bibitem{mundnich2020tiles}
K.~Mundnich, B.~M. Booth, M.~l’Hommedieu, T.~Feng, B.~Girault,
  J.~L’hommedieu, M.~Wildman, S.~Skaaden, A.~Nadarajan, J.~L. Villatte,
  et~al.
\newblock Tiles-2018, a longitudinal physiologic and behavioral data set of
  hospital workers.
\newblock {\em Scientific Data}, 7(1):1--26, 2020.

\bibitem{oneto2020fairness}
L.~Oneto and S.~Chiappa.
\newblock Fairness in machine learning.
\newblock In {\em Recent Trends in Learning From Data}, pages 155--196.
  Springer, 2020.

\bibitem{oneto2019taking}
L.~Oneto, M.~Doninini, A.~Elders, and M.~Pontil.
\newblock Taking advantage of multitask learning for fair classification.”
  the 2019 aaai.
\newblock In {\em ACM Conference (AIES)}, 2019.

\bibitem{panda2018contemplating}
R.~Panda, J.~Zhang, H.~Li, J.-Y. Lee, X.~Lu, and A.~K. Roy-Chowdhury.
\newblock Contemplating visual emotions: Understanding and overcoming dataset
  bias.
\newblock In {\em Proceedings of the European Conference on Computer Vision
  (ECCV)}, pages 579--595, 2018.

\bibitem{park2021comparison}
Y.~Park, J.~Hu, M.~Singh, I.~Sylla, I.~Dankwa-Mullan, E.~Koski, and A.~K. Das.
\newblock Comparison of methods to reduce bias from clinical prediction models
  of postpartum depression.
\newblock {\em JAMA network open}, 4(4):e213909--e213909, 2021.

\bibitem{rajkomar2018ensuring}
A.~Rajkomar, M.~Hardt, M.~D. Howell, G.~Corrado, and M.~H. Chin.
\newblock Ensuring fairness in machine learning to advance health equity.
\newblock {\em Annals of internal medicine}, 169(12):866--872, 2018.

\bibitem{rodolfa2021empirical}
K.~T. Rodolfa, H.~Lamba, and R.~Ghani.
\newblock Empirical observation of negligible fairness--accuracy trade-offs in
  machine learning for public policy.
\newblock {\em Nature Machine Intelligence}, 3(10):896--904, 2021.

\bibitem{owidmentalhealth}
H.~R. Saloni~Dattani and M.~Roser.
\newblock Mental health.
\newblock {\em Our World in Data}, 2021.
\newblock https://ourworldindata.org/mental-health.

\bibitem{simonyan2013deep}
K.~Simonyan, A.~Vedaldi, and A.~Zisserman.
\newblock Deep inside convolutional networks: Visualising image classification
  models and saliency maps.
\newblock {\em arXiv preprint arXiv:1312.6034}, 2013.

\bibitem{srivastava2019mathematical}
M.~Srivastava, H.~Heidari, and A.~Krause.
\newblock Mathematical notions vs. human perception of fairness: A descriptive
  approach to fairness for machine learning.
\newblock In {\em Proceedings of the 25th acm sigkdd international conference
  on knowledge discovery \& data mining}, pages 2459--2468, 2019.

\bibitem{su2014emotiono+}
Y.~Su, B.~Hu, L.~Xu, H.~Cai, P.~Moore, X.~Zhang, and J.~Chen.
\newblock Emotiono+: Physiological signals knowledge representation and emotion
  reasoning model for mental health monitoring.
\newblock In {\em 2014 IEEE International Conference on Bioinformatics and
  Biomedicine (BIBM)}, pages 529--535. IEEE, 2014.

\bibitem{swati2022early}
S.~Swati, M.~Kumar, and S.~Namasudra.
\newblock Early prediction of cognitive impairments using physiological signal
  for enhanced socioeconomic status.
\newblock {\em Information Processing \& Management}, 59(2):102845, 2022.

\bibitem{umetani1998twenty}
K.~Umetani, D.~H. Singer, R.~McCraty, and M.~Atkinson.
\newblock Twenty-four hour time domain heart rate variability and heart rate:
  relations to age and gender over nine decades.
\newblock {\em Journal of the American College of Cardiology}, 31(3):593--601,
  1998.

\bibitem{valdivia2021fair}
A.~Valdivia, J.~S{\'a}nchez-Monedero, and J.~Casillas.
\newblock How fair can we go in machine learning? assessing the boundaries of
  accuracy and fairness.
\newblock {\em International Journal of Intelligent Systems}, 36(4):1619--1643,
  2021.

\bibitem{vallor2017artificial}
S.~Vallor.
\newblock Artificial intelligence and public trust.
\newblock 2017.

\bibitem{wang2021hybrid}
D.~Wang, J.~Yu, L.~Chen, X.~Li, H.~Jiang, K.~Chen, M.~Zheng, and X.~Luo.
\newblock A hybrid framework for improving uncertainty quantification in deep
  learning-based qsar regression modeling.
\newblock {\em Journal of cheminformatics}, 13(1):1--17, 2021.

\bibitem{xia2018physiological}
L.~Xia, A.~S. Malik, and A.~R. Subhani.
\newblock A physiological signal-based method for early mental-stress
  detection.
\newblock {\em Biomedical Signal Processing and Control}, 46:18--32, 2018.

\bibitem{xu2020investigating}
T.~Xu, J.~White, S.~Kalkan, and H.~Gunes.
\newblock Investigating bias and fairness in facial expression recognition.
\newblock In {\em European Conference on Computer Vision}, pages 506--523.
  Springer, 2020.

\bibitem{yan2020mitigating}
S.~Yan, D.~Huang, and M.~Soleymani.
\newblock Mitigating biases in multimodal personality assessment.
\newblock In {\em Proceedings of the 2020 International Conference on
  Multimodal Interaction}, pages 361--369, 2020.

\bibitem{yu2020passive}
H.~Yu and A.~Sano.
\newblock Passive sensor data based future mood, health, and stress prediction:
  User adaptation using deep learning.
\newblock In {\em 2020 42nd Annual International Conference of the IEEE
  Engineering in Medicine \& Biology Society (EMBC)}, pages 5884--5887. IEEE,
  2020.

\bibitem{zanna2021clustering}
K.~Zanna, T.~Neal, and S.~Canavan.
\newblock Clustering of physiological signals by emotional state, race, and
  sex.
\newblock In {\em Companion Publication of the 2021 International Conference on
  Multimodal Interaction}, pages 312--316, 2021.

\bibitem{zhang2007effect}
J.~Zhang.
\newblock Effect of age and sex on heart rate variability in healthy subjects.
\newblock {\em Journal of manipulative and physiological therapeutics},
  30(5):374--379, 2007.

\bibitem{zhang2021survey}
Y.~Zhang and Q.~Yang.
\newblock A survey on multi-task learning.
\newblock {\em IEEE Transactions on Knowledge and Data Engineering}, 2021.

\bibitem{zvolensky2017disparities}
M.~J. Zvolensky, L.~Garey, and J.~Bakhshaie.
\newblock Disparities in anxiety and its disorders, 2017.

\end{thebibliography}

\end{document}